\documentclass[lettersize,journal]{IEEEtran}
\usepackage{amsmath,amsfonts,amssymb}
\usepackage{array}
\usepackage[caption=false,font=normalsize,labelfont=sf,textfont=sf]{subfig}
\usepackage{textcomp}
\usepackage{stfloats}
\usepackage{url}
\usepackage{verbatim}
\usepackage{graphicx}
\usepackage{cite}
\usepackage{multirow}
\usepackage{booktabs}
\usepackage{wrapfig}
\usepackage[table]{xcolor}
\usepackage{pifont}
\usepackage{makecell}
\usepackage{tikz}
\usepackage{bbm}
\usepackage{capt-of}
\usepackage{cuted}
\usepackage{needspace}
\usepackage[hidelinks]{hyperref}
\usepackage{orcidlink}

\definecolor{rowours}{gray}{0.92}
\definecolor{gray1}{gray}{0.85}
\definecolor{gray2}{gray}{0.90}
\definecolor{gray3}{gray}{0.95}

\newcommand{\eg}{e.g.}

\graphicspath{{figures/}}
\begin{document}

\title{Topology-Unified 2D Pose Estimation across Intact, Residual and Prosthetic Limbs}

\author{Tianye Qi\orcidlink{0009-0007-5770-0365}, Tengyue Zhang\orcidlink{0009-0001-5118-9208}, Jiaying Ying\orcidlink{0009-0003-8908-9231}, Tianqing Zhu\orcidlink{0000-0003-3411-7947}, and Xin Yu\orcidlink{0000-0002-0269-5649}
\thanks{Tianye Qi and Jiaying Ying are with The University of Queensland (e-mail: uqtqi4@uq.edu.au).}
\thanks{Tengyue Zhang and Xin Yu are with Australian Institute for Machine Learning, Adelaide University.}
\thanks{Tianqing Zhu is with City University of Macau.}}

\markboth{Preprint -- Under Review}{Qi \MakeLowercase{\textit{et al.}}: Topology-Unified 2D Pose Estimation}

\maketitle

\setlength{\stripsep}{0pt}
\begin{strip}
  \vspace*{-33pt}
  \centering
  \includegraphics[width=\textwidth]{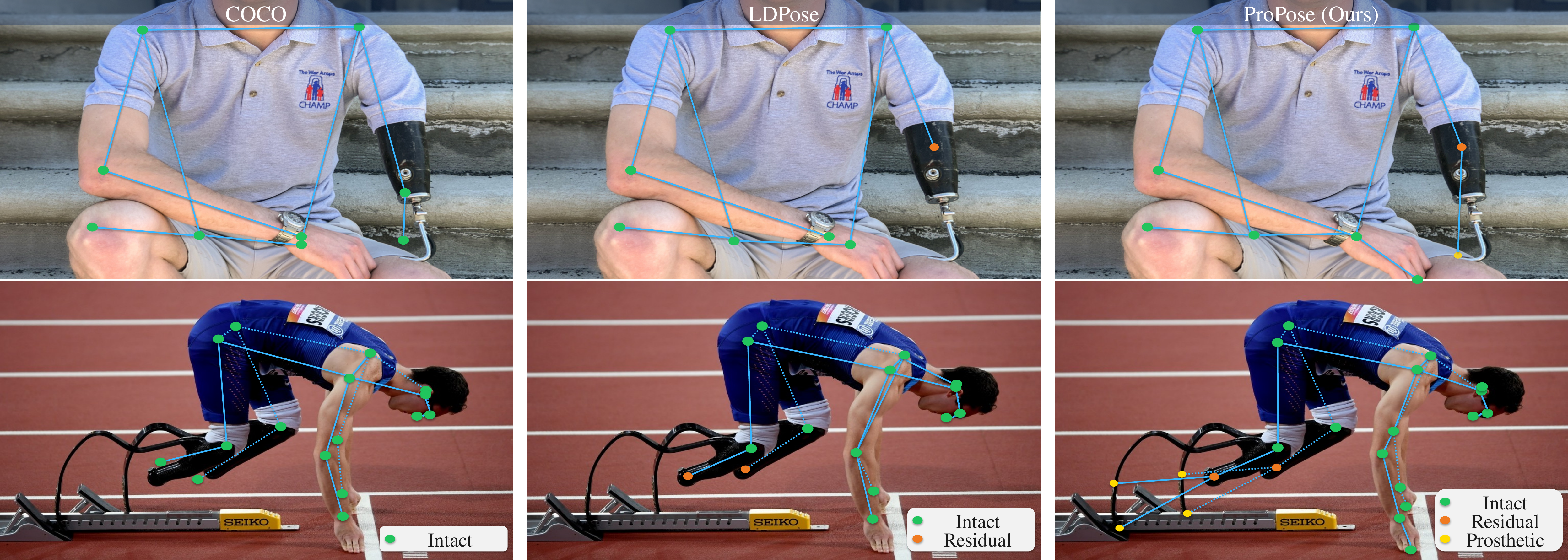}
  \captionof{figure}{Resolving topological ambiguity, on an upper limb (top) and a lower limb (bottom). Unlike COCO~\cite{lin2014microsoft} that hallucinates biological joints on mechanical structures, and LDPose~\cite{ying2025ldpose} that truncates topology at the residual limb, ProPose explicitly disentangles joint semantics to accurately reconstruct the complete kinematic chain across intact, residual, and prosthetic limbs. In the top row the elbow and wrist of the prosthetic arm do not exist and carry no coordinate: COCO has no way to record that and must place them on the device, LDPose stops at the residual limb, and only ProPose ends the chain where the limb ends and resumes it at the prosthetic fingertip.}
  \label{fig:protocol comparison}
  \vspace{15pt}
\end{strip}

\begin{abstract}
Driven by the availability of large-scale datasets, Human Pose Estimation (HPE) plays a critical role in numerous downstream tasks. However, mainstream benchmarks exhibit severe representation bias, predominantly featuring able-bodied individuals. While a few pioneering datasets have attempted to address limb differences, their annotation protocols fail to generalize, struggling to represent specialized mechanical structures like running blades or unprosthetized residual limbs. To bridge this gap, we introduce \textbf{ProPose}, a large-scale benchmark featuring a novel annotation protocol that unifies the topological representation of biological limbs, diverse prostheses, and physical absences within a single framework. Because real-world prosthetic images are inherently scarce and exhibit extreme long-tail distributions, we design a \textbf{Real-to-Synthetic data expansion pipeline} to explicitly synthesize and expand the underrepresented cases. 
However, simply training existing models on this enriched dataset often leads to suboptimal solutions, as they estimate each keypoint independently and might hallucinate non-existent joints on mechanical structures. To resolve this, we propose ProLoss, a structure-aware objective that enforces keypoint dependencies within a single limb to prevent unrealistic limb predictions. Extensive experiments demonstrate that our approach improves the classification accuracy of long-tail prosthetic joints by 2\% to 6\% without compromising spatial coordinate localization performance. This work sets a foundation for inclusive pose estimation, unlocking new possibilities for understanding the interactions between human bodies and assistive devices.
\end{abstract}

\begin{IEEEkeywords}
Pose estimation with Prostheses \and Body Structure-Aware Representation \and Prosthesis-aware Pose Estimation Benchmark\end{IEEEkeywords}

\Needspace*{4\baselineskip}
\section{Introduction}

\IEEEPARstart{H}{uman} Pose Estimation (HPE) stands as a pivotal task in computer vision, serving as the cornerstone for a myriad of downstream applications. Its precise localization of anatomical keypoints is indispensable for action recognition \cite{yan2018spatial, duan2022revisiting}, physical literacy analysis \cite{guo2025plnet}, 3D human mesh recovery \cite{cai2023smpler, goel2023humans, wang2025blade}, and clinical gait analysis \cite{uhlrich2023opencap, stenumclinical}. Driven by the rapid advancement of deep learning, recent years have witnessed the emergence of robust state-of-the-art frameworks. Representative methods include transformer-based ViTPose \cite{xu2022vitpose}, and real-time RTMPose \cite{jiang2023rtmpose} and YOLO-Pose series \cite{maji2022yolo,tian2025yolov12,sapkota2025yolo26}. The impressive performance of these models is largely attributed to the availability of large-scale, high-quality benchmarks. Beyond foundational datasets, like MS COCO \cite{lin2014microsoft} and MPII \cite{andriluka14cvpr}, the field has advanced through benchmarks focusing on complex occlusion (CrowdPose \cite{li2019crowdpose}, OCHuman \cite{zhang2019pose2seg}) and fine-grained whole-body perception (COCO-WholeBody \cite{jin2020whole}).

Despite their scale, these mainstream benchmarks suffer from a significant representation bias, predominantly featuring able-bodied individuals. A few pioneering works have attempted to bridge this gap, though they adopt different annotation protocols. LDPose \cite{ying2025ldpose} focuses on annotating residual limbs, which are highly valuable for studying the amputation site, but it leaves prosthetic devices unannotated. ProGait \cite{yin2025progait} includes the prosthesis by mapping mechanical components to standard biological keypoints (\emph{e.g.}, knee, ankle). While this strategy aligns well with anthropomorphic prostheses that mimic human geometry, biological mapping can introduce semantic ambiguities when applied to specialized prostheses with non-biological geometries, such as running blades, or to exposed residual limbs without prosthetic attachment. Consequently, it is highly desirable for a unified pose annotation protocol that can provide topological consistency across both biological limbs and diverse mechanical structures.  

To bridge this critical gap, we introduce \textbf{ProPose} \footnote{Code and dataset are available at \url{https://github.com/SoraLink/ProPose} and \url{https://huggingface.co/datasets/SoraLink/ProPose}.}, a large-scale benchmark specifically designed to resolve the representation bias and topological mismatch in prosthesis-aware pose estimation (see Fig.~\ref{fig:protocol comparison}). Advancing beyond the annotation scheme of LDPose \cite{ying2025ldpose}, which only appends eight residual points while neglecting distal interactions, we propose an \textbf{Omni-Pose Protocol} that integrates two key innovations: 
First, to unify the topological representation across diverse prosthetic designs, we explicitly label distal extremities (fingertips, heels, and toes). This ensures that regardless of the specific mechanical configuration, a coherent kinematic chain and effective contact points are consistently captured within a single framework. Second, to endow the model with the ability to distinguish between biological limbs and mechanical devices, and to handle cases where certain joints are physically non-existent (\emph{e.g.}, specialized prostheses or residual limbs without prosthetic devices), we introduce a novel \textbf{semantic attribute (Keypoint Type)} for each joint, categorizing them as \textit{Biological}, \textit{Prosthetic}, or \textit{Absent}.

After annotating the semantic labels and prosthetic keypoints on LDPose dataset~\cite{ying2025ldpose}, we found that the number of prosthetic keypoints is very scarce, especially for elbow and knee joints. Because the number of images of individuals with prostheses are much smaller and LDPose already includes many of the available sources, our initial web crawling efforts yielded only 1,807 new images. To overcome this critical data bottleneck, we designed a \textbf{Real-to-Synthetic data expansion pipeline} that leverages Nano Banana~\cite{comanici2025gemini} for data augmentation. We reserve 437 high-quality real-world images (covering 499 unseen subjects) from this crawled subset and use them for testing. The remaining 1,370 images are processed through the pipeline for image synthesis, followed by manual review to remove generated samples containing visible artifacts. As a result, we expand our training set with 9,558 images (containing 10,345 instances).

The data expansion significantly mitigates the long-tail distribution observed in prior benchmarks. For instance, prosthetic elbows were severely underrepresented in LDPose, accounting for only 0.19\% of all annotated elbow keypoints; following our augmentation pipeline, this proportion rises to 5.81\%, a 31-fold increase. The proportion of prosthetic wrists has grown 8.6-fold as well, \emph{i.e.}, from 1.00\% to 8.63\%.
Therefore, our ProPose dataset ensures that models can learn robust features across a wide spectrum of prosthetic devices, rather than overfitting to biological limbs.

Furthermore, directly employing standard training objectives will lead to anatomically contradictory predictions, such as hallucinating a biological wrist from a residual limb, because they treat keypoints as independent spatial entities. In practice, human-prosthetic anatomy is governed by strict structural and semantic constraints, in which one joint determines the physical validity of the others on the same limb. To eliminate unreasonable limb configurations, we propose \textbf{ProLoss}, a structure-aware limb-dependence function. By explicitly modeling the semantic dependencies across all joints within a single limb, ProLoss penalizes invalid type combinations, forcing a network to output logically consistent kinematic chains. Extensive experiments demonstrate that the models trained with ProLoss on our ProPose dataset achieve superior results across biological, residual and prosthetic limbs. This work effectively establishes topological consistency across diverse limb types, providing a unified framework for inclusive human pose estimation. In summary, our contributions are threefold:
\begin{itemize}
    \setlength\itemsep{0.15em}
    
    \item \textbf{Unified Pose Representation Protocol:} We propose a unified 2D human pose representation protocol, namely \textbf{Omni-Pose}, for diverse limb configurations. Particularly, \textbf{Omni-Pose} involves a semantic attribute (Keypoint Type) for each joint to determine limb types.

    \item \textbf{Large-Scale Benchmark (ProPose):} We construct \textbf{ProPose} dataset that not only covers normal and residual limbs but also includes prosthetic limbs. Moreover, to overcome data scarcity and long-tail distributions of prosthesis images, we design a \textbf{Real-to-Synthetic data expansion pipeline} to enrich the quantity and diversity of prosthetic devices.

    \item \textbf{Structure-Aware Objective (ProLoss):} We propose \textbf{ProLoss}, a structure- and semantics-aware limb-dependent loss that explicitly models intra-limb semantic dependencies. It penalizes unreasonable limb configurations, forcing the network to learn physically plausible kinematic chains.

\end{itemize}

\section{Related Work}
\subsection{Human Pose Estimation Benchmarks}
\paragraph{Full-limb Benchmarks.}
The rapid advancement of 2D Human Pose Estimation (HPE) has been fundamentally driven by large-scale annotated benchmarks. Foundational datasets such as MS COCO \cite{lin2014microsoft} and MPII \cite{andriluka14cvpr} establish the standard for multi-person pose estimation, fueling the development of diverse data-driven architectures. To address more challenging scenarios characterized by severe occlusion and crowded scenes, CrowdPose \cite{li2019crowdpose} and OCHuman \cite{zhang2019pose2seg} are introduced. Furthermore, benchmarks have been extended to the temporal domain with PoseTrack21 \cite{doering22} for video-based tracking, and to 3D space with Human3.6M \cite{Zhu_2023_ICCV} and FreeMan \cite{wang2024freeman}. However, these benchmarks and models exhibit a systemic bias: they implicitly assume a canonical able-bodied topology (\emph{e.g.}, the standard 17-keypoint format). As a result, they often hallucinate missing limbs or fail to recognize prosthetic devices.

\paragraph{Specialized Benchmarks.}
To bridge the representation gap, specialized benchmarks have recently emerged. \textbf{LDPose} \cite{ying2025ldpose} pioneers this direction by focusing on limb deficiency. It provides 28K images and extends the standard COCO topology by appending 8 keypoints to localize residual limb endpoints. Although LDPose acknowledges the presence of prosthetics, it \textit{excludes them from annotations}, limiting the model's ability to analyze the interaction between humans and assistive devices. \textbf{Inclusive VidPose} \cite{du2026inclusivevidpose} extends the LDPose to the video domain with 313 clips totaling 327k frames. It leverages temporal context to effectively disambiguate occluded limbs from truly absent ones, but it still lacks keypoint annotations for prosthetic devices. \textbf{ProGait} \cite{yin2025progait} introduces a video benchmark (412 clips) that explicitly labels prosthetic joints. It maps prostheses to biological joint labels, but is constrained by its small scale and limited diversity of prosthetic devices.
Thus, models trained on ProGait fail to accommodate specialized devices (\emph{e.g.}, running blades) or distinguish prostheses from \textit{intact limbs}. 
These limitations underscore the necessity for a large-scale benchmark that features diverse prosthetic configurations and explicitly encodes the \textbf{semantic attribute} of each joint. 

\subsection{Human Pose Estimation (HPE) Methods}

HPE algorithms can often be grouped into two categories: top-down and bottom-up methods \cite{liu2022recent}. \textbf{Top-down} approaches first detect bounding boxes and then estimate keypoints.  Representative large models include {HRNet} \cite{wang2020deep} and the transformer-based {ViTPose} \cite{xu2022vitpose}. However, the computational cost of top-down architectures scales linearly with the number of instances. To address this, real-time top-down models like {RTMPose} \cite{jiang2023rtmpose} and distillation-based {DWPose} \cite{yang2023effective} are proposed to balance efficiency and accuracy via optimized backbones and lightweight heads.

\textbf{Bottom-up} or single-stage methods prioritize inference speed by eliminating the separate cropping step. {OpenPose} \cite{8765346} pioneers the bottom-up paradigm using Part Affinity Fields (PAFs) to efficiently associate body parts for multi-person estimation. Following this direction,  models like HigherHRNet \cite{cheng2020higherhrnet} address scale variation issues using high-resolution feature pyramids, while {DEKR} \cite{GengSXZW21} utilizes adaptive convolutions for direct keypoint regression. More recently, {YOLO-Pose} \cite{maji2022yolo} embeds pose estimation heads directly into a single-stage object detector, offering end-to-end inference.
Since these models are also trained on images of able-bodied individuals, they struggle recognizing prosthetic devices or hallucinate missing limbs. 

\subsection{Data Augmentation and Synthesis}

Acquiring large-scale, annotated datasets for limb-deficient individuals is inherently challenging due to the underrepresented nature of the population. To address this low-resource data challenge, synthetic data generation has been widely adopted~\cite{cao2025analytical}. Traditional approaches rely on 3D rendering engines to create diverse poses and environments. For instance, {SURREAL} \cite{varol2017learning} generates synthetic humans from 3D scans. More relevant to our domain, {WheelPose} \cite{huang2024wheelpose} utilizes a synthetic engine to generate data for wheelchair users, effectively addressing severe occlusions caused by assistive equipment. However, these engine-based methods often suffer from a significant \textit{domain gap} between synthetic renders and photorealistic imagery, limiting their generalization to real-world scenarios.

Recent advances in MLLMs have increasingly focused on human-centric perception and understanding, including identity association, body pose, action, and interaction reasoning \cite{guo2026beyond, guo2026ssmnbenchdiagnosingimagebasedcrossview}. These advances have facilitated instruction-based human image editing, enabling models to follow textual commands while preserving identity and plausibly modifying pose, limb configuration, and appearance. Unlike 3D rendering pipelines that generate images from scratch, state-of-the-art models such as Qwen-Image-Edit \cite{wu2025qwenimagetechnicalreport} and Gemini \cite{comanici2025gemini} demonstrate the ability to modify existing real-world images through precise natural language prompts. While text-to-image synthesis has also advanced considerably through latent diffusion models \cite{rombach2022high} and ControlNet \cite{zhang2023adding}, these methods often struggle to preserve the photorealism and subject identity when performing complex structural transformations, such as altering a posture from standing to running. In contrast, MLLMs excel at preserving these critical attributes and generating diverse, high-fidelity training samples directly from real-world data.

\section{Construction of the Proposed ProPose Dataset}

We introduce the procedure of data collection, annotation and post-processing in detail, and then present key statistics of ProPose dataset. As indicated in Table~\ref{dataset_comparison}, we also compare ProPose with prior datasets.

\begin{table*}[t]
    \centering
    \footnotesize
    \caption{Comparison of human pose datasets. While recent efforts have introduced datasets for specific disabilities, none of the existing benchmarks unifies intact (Int), residual (Res), and prosthetic (Pros) limbs.}
    \resizebox{\linewidth}{!}{  
    \scriptsize 
    \begin{tabular}{lcccccccc}
        \toprule
        {Dataset} & {Total} & {Total} & {Avg. Poses} & {Limb} & {Resolution} & {Annotation} & {Env.} \\
        & {Images/Videos} & {Instances} & {per Image} & {Types} & & {Source} & {Complexity} \\
        \midrule

        MPII \cite{andriluka14cvpr} & 25k & 40k & $\sim$2 & Int & 1920 × 1080 & Crowd-sourced & Moderate \\
        MSCOCO \cite{lin2014microsoft} & 200k & 250k & $\sim$2 & Int & 640 × 480 to 1280 × 720 & Crowd-sourced & High \\
        CrowdPose \cite{li2019crowdpose} & 20k & 80k & $\sim$4 & Int & 1080 × 720 & Crowd-sourced & High \\
        OCHuman \cite{zhang2019pose2seg} & 4.7k & 8k & $\sim$2 & Int & 1280 × 720 & Crowd-sourced & High \\
        PoseTrack21 \cite{doering22} & 66k & 177k & $\sim$2.5 & Int  & 1280 × 720 & Crowd-sourced & High \\
        FreeMan \cite{wang2024freeman} & 11.3m & 11.3m & $\sim$1 & Int & 1920 × 1080 & Crowd-sourced & High \\
        Human3.6M \cite{Zhu_2023_ICCV} & 3.6m & 3.6m & $\sim$1 & Int & 1920 × 1080 & Experts & Low \\
        LDPose \cite{ying2025ldpose} & 28k & 72.7k & $\sim$3 & Int, Res & 200 × 143 to 8100 × 4752 & Medical Experts & High \\
        ProGait \cite{yin2025progait} & 412 & 412 & $\sim$1 & Int, Pros & 1920 × 1080 & Medical Experts & Moderate \\
        WheelPose \cite{huang2024wheelpose} & 2.5k & 2.6k &$\sim$1 & Int & 480 × 360 to 1280 × 1280 & Generated & Moderate \\
        \textbf{ProPose(Ours)}       & 36.4k & 63.0k & $\sim$2 & Int, Res, Pros  & 200 × 143 to 8640 × 8256 & Medical Experts & High\\
       
        \bottomrule
    \end{tabular}
    }
    \label{dataset_comparison}
\end{table*}

\subsection{Omni-Pose Representation Protocol}

\label{sec:protocol}

Standard pose estimation protocols (\emph{e.g.}, COCO-17) are designed for able-bodied individuals. LDPose \cite{ying2025ldpose} appended 8 residual keypoints for residual limbs but neglects the existence of prosthetic devices. Although ProGait aligns prosthetic devices with the standard MS COCO topology, this biological mapping fails to capture scenarios in which specific joints are structurally absent in certain prostheses. For instance, a carbon-fiber running prosthesis lacks an anatomical ankle. Consequently, current annotation protocols fall short in their ability to unify all limb types, including intact, residual, and prosthetic limbs.

\begin{figure}[!t]
  \centering
  \includegraphics[width=\linewidth]{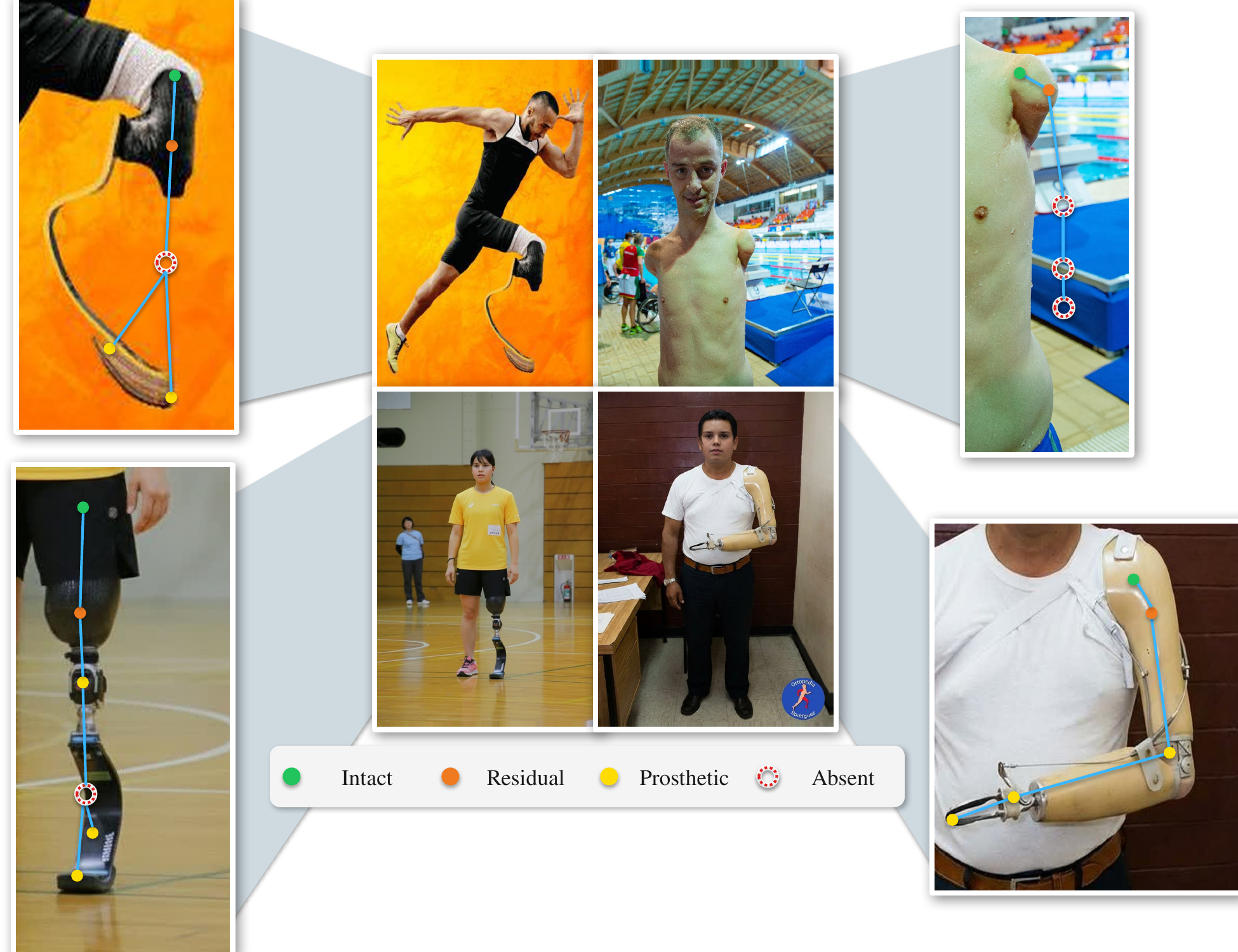}
  \caption{Omni-Pose protocol. Omni-Pose unifies intact, residual, prosthetic and absent limb configurations within a single topology; each corner magnifies the limb of the photograph it points to. Filled markers are annotations taken from the dataset. Absent joints carry no coordinate by definition, so they are drawn as open dashed rings at their inferred position for illustration only.}
  \label{fig:proposed_topology}
\end{figure}
In this work, we generalize the LDPose protocol by explicitly adding 6 distal keypoints: \textit{Heel}, \textit{Toe}, and \textit{Fingertip} (Left/Right), as well as assign a \textbf{Semantic Type} to each keypoint to faithfully reflect the physical conditions observed in the images (see Fig.~\ref{fig:proposed_topology}). Specifically, (1) \textbf{Limb Joints} (including standard limb joints and the newly added distal points) can appear in all three types: $c \in \{\textit{Biological}, \textit{Prosthetic}, \textit{Absent}\}$; (2) \textbf{Residual Keypoints}, representing the physical stump, only exist as $c \in \{\textit{Biological}, \textit{Absent}\}$; and (3) \textbf{Other Core Joints} (\eg, head and torso) are universally \textit{Biological}. This categorization captures prosthetic replacements and missing limb structures.

\textit{Visibility Protocol:}
Conventional visibility labels are designed to describe the observability of keypoint coordinates, typically distinguishing visible keypoints, occluded but inferable keypoints, and uncertain keypoints. However, these labels do not directly capture anatomical absence, where no valid coordinate exists but the absence of a joint can be determined. To address this limitation, we introduce an ``Absent'' semantic type under a deterministic visibility protocol. For example, when a residual limb is clearly visible and no prosthesis is present, the absence of subsequent joints is an observable anatomical state. We therefore assign these keypoints a visible visibility flag ($v=2$) together with the ``Absent'' semantic type, indicating that the absence itself is visible and deterministic, even though no coordinate is annotated.

\subsection{Real-to-Synthetic Data Expansion Pipeline}
\label{sec:pipeline}

Initial statistical analysis of the LDPose training dataset \cite{ying2025ldpose} reveals a severe imbalance: instances containing prosthetic elbows constituted approximately 0.2\% of the total annotations. Our preliminary experiments show this scarcity leads to severe model bias after fine-tuning. However, due to the long-tail nature of prosthetic imagery, traditional web crawling alone cannot fully resolve this problem. Thus, we propose a real-to-synthetic data expansion pipeline, which comprises three phases: real-world collection, generative AI-driven data augmentation, and quality control. 
The pipeline can effectively address data scarcity while ensuring that the synthesized data remain realistic and useful for model training.

\paragraph{Phase 1: Data Collection.}
We collect images from social media and search engines. After de-duplication and manual screening, we obtain 1,807 high-quality real-world images. We use 1,370 images as seeds for training data generation, and integrate the remaining 437 into the LDPose test set for evaluation.

\paragraph{Phase 2: Generative AI-driven Image-to-Image Synthesis.}
Even with the addition of web-crawled images, the proportion of images with prosthesis is small. To enrich the training set, we employ an instruction-driven editing pipeline using Nano Banana. Here, we adopt a multi-stage strategy:

\begin{enumerate}
    \setlength\itemsep{0.25em}
    \item \textbf{Contextual Outpainting:} Many crawled images feature close-up shots with truncated limbs, lacking global context. We first use the model (\emph{i.e.}, Nano Banana) to outpaint these crops into half-body or full-body images for robust pose estimation.
    \item \textbf{Pose Editing and Balancing:} We alter the subjects' poses using detailed text prompts. To prevent the model from generating physically impossible structures, we explicitly specify the valid bending joints and the rigid, non-deformable components of the prosthetic devices. Additionally, we perform targeted oversampling, specifically generating samples containing rare prosthetic categories (\emph{e.g.}, prosthetic elbows and knees) to alleviate the long-tail distributions issue.
\end{enumerate}

Before finalizing the image editor used in this phase, we conduct a model-selection study to determine which instruction-based editing model is suitable for prosthesis-aware data expansion. For each candidate model, we manually review 100 generated outputs using three task-specific criteria: whether the human pose is clearly edited, whether the original prosthetic or residual-limb condition is preserved, and whether severe visual artifacts are present. Table~\ref{tab:gen_model_qc} summarizes this quality-control comparison.

The Gemini-based models achieve the strongest overall quality. Gemini Flash obtains the highest acceptance rate (80.0\%), while Gemini 3 Pro also maintains a high acceptance rate (72.0\%) with all reviewed samples showing clear pose edits. In contrast, the non-Gemini models have acceptance rates of 36.0\% or lower. GPT Image 2 preserves the prosthetic device when the edit succeeds, but only 36.0\% of its outputs show clear pose edits. Seedream, Qwen-Image-Edit, and FireRed-Image-Edit frequently fail to preserve prosthetic structures or introduce severe artifacts. We therefore choose the Gemini family for data extension, because it provides the best balance between controllable pose editing, prosthesis preservation, and low artifact rate.

\begin{table*}[!t]
\centering
\caption{Quality-control comparison of generative models for prosthesis-aware data expansion. For each model, we generate 100 candidate images and manually screen them for image quality. A candidate is accepted only when the pose is clearly edited, the prosthetic or residual-limb condition is preserved, and no severe visual artifact is present.}
\label{tab:gen_model_qc}
\scriptsize
\setlength{\tabcolsep}{4pt}
\resizebox{\textwidth}{!}{
\begin{tabular}{lcccccc}
\toprule
\textbf{Generative model} & \textbf{Generated} & \textbf{Accepted}~$\uparrow$ & \textbf{Acceptance rate}~$\uparrow$ & \textbf{Pose edited}~$\uparrow$ & \textbf{Prosthesis preserved}~$\uparrow$ & \textbf{Artifact rate}~$\downarrow$ \\
\midrule
Gemini Flash Image / Nano Banana 2 & 100 & 80 & \textbf{80.0\%} & \textbf{100.0\%} & \textbf{96.0\%} & 4.0\% \\
Gemini 3 Pro Image / Nano Banana Pro & 100 & 72 & 72.0\% & \textbf{100.0\%} & 88.0\% & 12.0\% \\
GPT Image 2 & 100 & 36 & 36.0\% & 36.0\% & \textbf{100.0\%} & 8.0\% \\
Seedream / Jimeng & 100 & 16 & 16.0\% & 36.0\% & 56.0\% & 4.0\% \\
Qwen-Image-Edit & 100 & 12 & 12.0\% & 28.0\% & 28.0\% & 52.0\% \\
FireRed-Image-Edit & 100 & 8 & 8.0\% & 20.0\% & 20.0\% & 72.0\% \\
\bottomrule
\end{tabular}
}
\end{table*}

\begin{figure*}[!t] 
    \centering
    \begin{minipage}[c]{0.42\textwidth} 
        \centering
        \includegraphics[width=\textwidth]{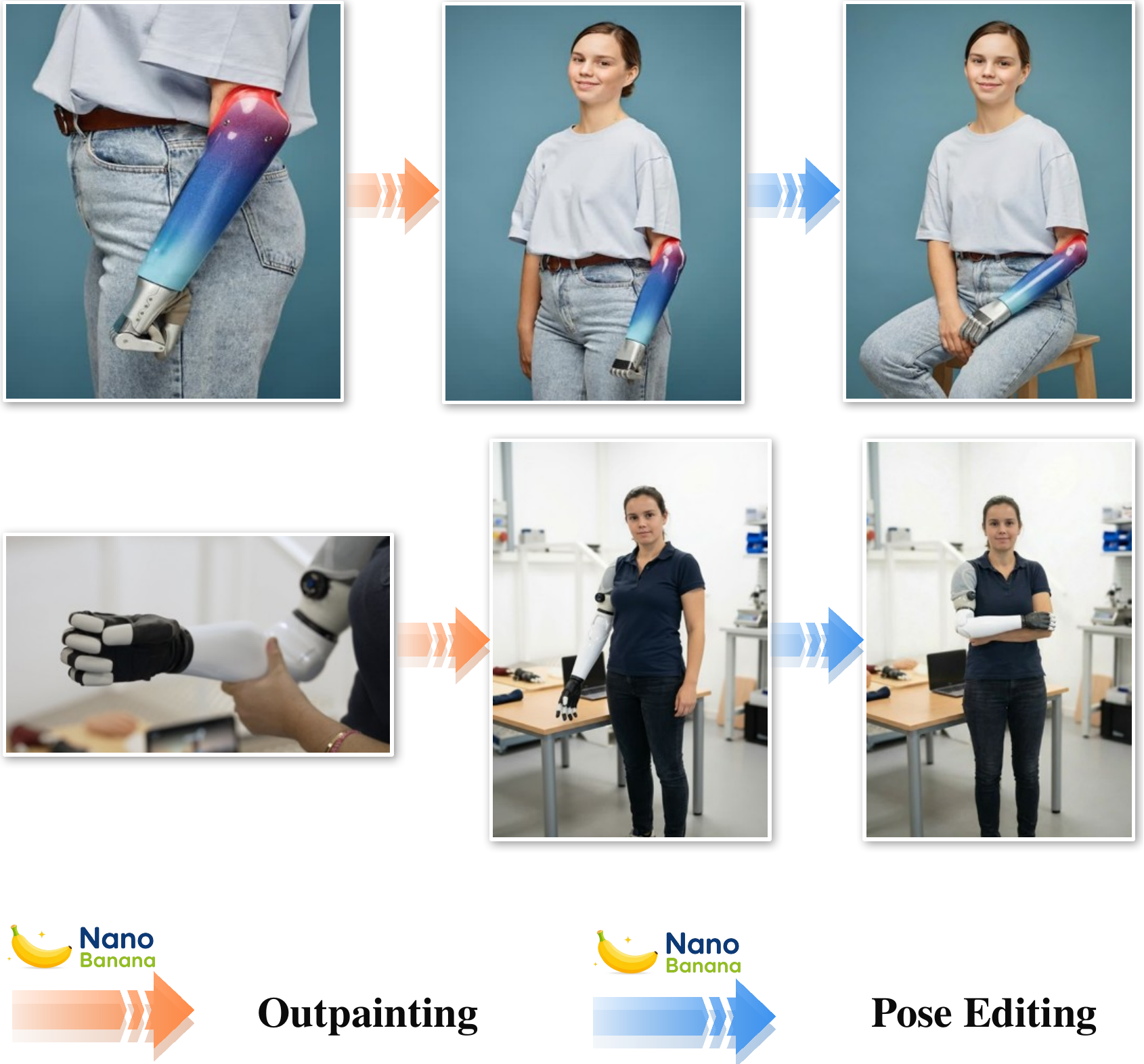}
        \label{fig:image_a}
    \end{minipage}%
    \hfill
    \begin{minipage}[c]{0.01\textwidth}
        \centering
        \begin{tikzpicture}
            \draw[dash pattern=on 3pt off 2pt, thick, gray] (0,-3.6) -- (0, 3.2);
        \end{tikzpicture}
    \end{minipage}%
    \hfill
    \begin{minipage}[c]{0.56\textwidth} 
        \centering
        \begin{minipage}[c]{\textwidth}
            \centering
            \includegraphics[width=\textwidth]{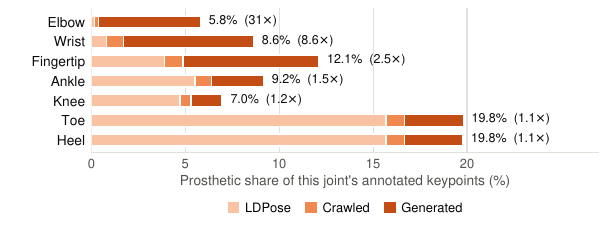}
            \label{fig:image_b}
        \end{minipage}

        \begin{minipage}[t]{0.49\textwidth}
            \centering
            \includegraphics[width=\textwidth]{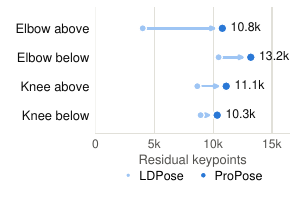}
            \label{fig:joint_analysis_c}
        \end{minipage}
        \hfill
        \begin{minipage}[t]{0.49\textwidth}
            \centering
            \includegraphics[width=\textwidth]{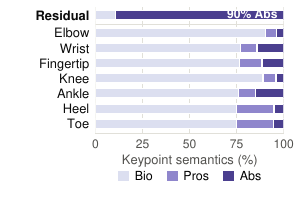}
            \label{fig:joint_analysis_d}
        \end{minipage}
    \end{minipage}

    \caption{Proposed generative AI-driven synthesis and dataset analysis. Statistics in (b)--(d) are computed over the training and test splits together, counting only annotated keypoints (visibility $>0$). (a) Outpainting and pose editing for diverse joint types. (b) Prosthetic keypoints as a share of each joint's annotations, split by the data source that contributed them; the multiplier in brackets is the growth of that share relative to the LDPose image subset alone. (c) Residual keypoints annotated as physically present, in the LDPose subset and in ProPose. (d) Composition of each row's annotated keypoints by semantic type; each row is normalised independently, and the separated \texttt{Residual} row aggregates the eight residual keypoints, which admit only \texttt{Bio} and \texttt{Abs}.}
    \label{fig:combinFed_image_and_data_distribution}
\end{figure*}

\paragraph{Phase 3: Quality and Privacy Assurance.}
To ensure data quality and adhere to ethical codes of conduct, we implement a rigorous filtering protocol:
\begin{itemize}
    \setlength\itemsep{0em}
    \item \textbf{Cascade De-duplication:} To prevent data leakage and ensure sample uniqueness, we apply a multi-stage filtering strategy. We first utilize {MD5 hashing} to automatically eliminate exact duplicates. Subsequently, we employ {Perceptual Hashing (pHash)} with a Hamming distance threshold of 5 to cluster structurally similar images. These candidate clusters are then manually reviewed to eliminate redundant near-duplicate images \cite{zauner2010implementation, schuhmann2022laion}.
    
    \item \textbf{Dual-Verification Protocol (Realism \& Kinematics):} Three annotators inspect each generated image, discarding those with visual artifacts or kinematic implausibility (\eg, impossible joint angles). This rigorous process removed approximately 10,000 images, ensuring the retained data strictly satisfies real-world biomechanical constraints.

    \item \textbf{Annotation Quality Verification:} 
    All images are independently annotated by two biomechanics experts. 
    When coordinate or semantic type annotations are inconsistent, a third expert reviews and resolves the annotations. 
    Overall, 94.0\% of samples reach direct agreement, 5.8\% are resolved after third-expert review, and 0.2\% are discarded due to unresolved annotation ambiguity.
    
    \item \textbf{Data Pruning of LDPose images:} We found that LDPose contains many able-bodied instances at very low resolution or with heavy occlusion. To prevent these samples from diluting the evaluation focus, we filter out low-resolution able-bodied samples, comprising around 20\% of the original training and test sets. Together with the pruned LDPose dataset, our additionally crawled and generated images form our ProPose dataset.
    
    \item \textbf{Copyright Compliance and De-identification:} We ensure compliance with the copyright and usage rights of the original images. For images with unclear copyright or licensing status, we implement a de-identification protocol using the \textit{FaceFusion} framework. Specifically, an identity-agnostic face swapping method \cite{chen2020simswap} is employed to replace original faces with a synthetic non-existent identity. This process removes original biometric features while preserving pose information.
\end{itemize}

To further support reproducibility, we provide additional details of our Real-to-Synthetic data expansion pipeline. Following the model-selection results in Table~\ref{tab:gen_model_qc}, we use the Gemini-based Nano Banana API, specifically \texttt{gemini-3-pro-image-preview}, for instruction-based image editing in the final data construction pipeline. During quality control, approximately 62\% of the generated images are discarded due to visual artifacts, implausible prosthetic structures, or inconsistent body configurations, and 9,558 generated images are retained for the final training set. No generated image is placed in the test split, so evaluation is performed entirely on real photographs.

\begin{figure*}[!t]
    \centering
    \includegraphics[width=\textwidth]{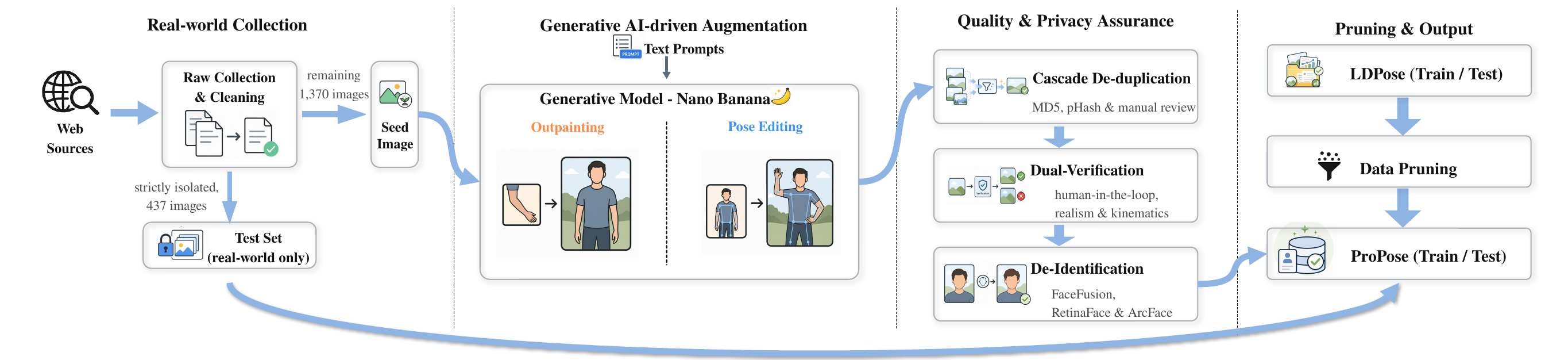}
    \caption{Overview of the Real-to-Synthetic data expansion pipeline. Phase~1 collects and de-duplicates web imagery and sets aside a strictly isolated real-world test set. Phase~2 uses instruction-driven editing to outpaint truncated crops into full-body images and to edit poses under kinematic constraints, with targeted oversampling of rare prosthetic categories. Phase~3 filters the generated images by cascade de-duplication, dual verification against realism and kinematics, and de-identification. Pruned LDPose images and the verified synthetic images together form the ProPose training set; the test set contains real photographs only.}
    \label{fig:supp_pipeline}
\end{figure*}

\begin{figure*}[!t]
    \centering
    \includegraphics[width=\textwidth]{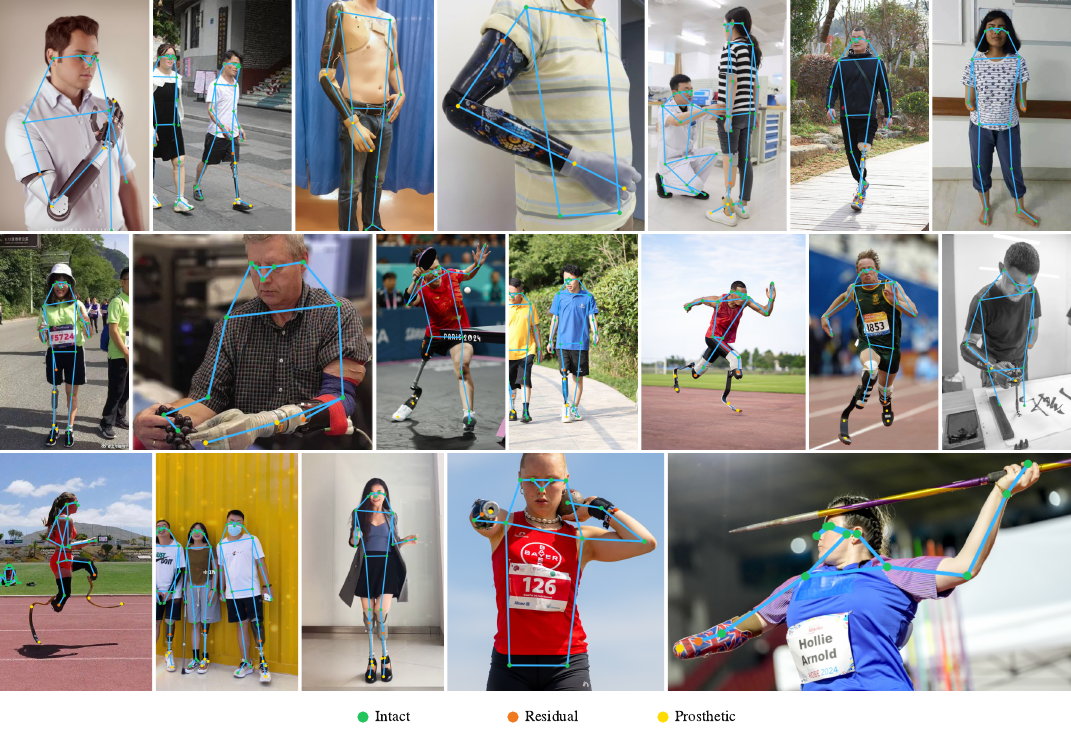}
    \caption{The Omni-Pose protocol applied across the ProPose dataset. The examples span the limb configurations the protocol has to describe: upper and lower limb, amputation above and below the joint, one side and both, and limbs with and without a prosthetic device. A keypoint is coloured by its semantic type, so a single skeleton can mix all three: \textit{Intact} covers both healthy and residual biological joints, \textit{Prosthetic} marks a joint realised by a mechanical device, and a joint labelled \textit{Absent} carries no coordinate at all, which is why a kinematic chain simply ends where the limb does. The last two examples show the attribute is assigned per joint rather than per limb: the biological wrist is \textit{Absent} while the fingertips of the attached prosthetic hand are \textit{Prosthetic}.}
    \label{fig:protocol_demo}
\end{figure*}

\subsection{Dataset Statistics of Our ProPose Dataset}
\label{sec:stats}

The statistical distribution of \textbf{ProPose} demonstrates two critical advancements: mitigation of long-tail scarcity and fine-grained semantic categorization. 

\paragraph{Mitigation of Long-Tail Data Scarcity.} 
Our data curation reshapes the distribution of limb-deficient data. As illustrated in Fig.~\ref{fig:combinFed_image_and_data_distribution} (b), our data construction successfully boosts rare categories across training and test sets. 
Relative to the LDPose image subset annotated under the same protocol, the prosthetic share of elbow keypoints rises from 0.19\% to 5.81\% (a 31$\times$ increase) and that of wrists from 1.00\% to 8.63\% (8.6$\times$), while the already better-represented lower-limb joints change by at most 1.5$\times$. The expansion is therefore concentrated exactly where the tail was thinnest. Because the original test set likewise lacked sufficient samples to reliably evaluate these rare joints, we expand it as well: prosthetic elbow and wrist keypoints grow from 38 to 129 and from 135 to 457, respectively.

Furthermore, Fig.~\ref{fig:combinFed_image_and_data_distribution} (c) quantifies the absolute increase in residual annotations. Compared to the LDPose subset, ProPose adds at least 1,400 annotated residual keypoints at every amputation level, and 6,770 at the above-elbow level, which was previously the scarcest. This expansion offers richer signals for learning complex geometries and robust evaluation.

\paragraph{Distributions of Keypoint Semantics.} 
ProPose explicitly labels each keypoint as Biological, Prosthetic, or Absent. Despite our data expansion, Fig.~\ref{fig:combinFed_image_and_data_distribution} (d) reveals that a severe long-tail distribution persists across specific semantic states. For instance, prosthetic elbows and absent heels remain rare minorities compared to dominant biological joints. This semantic imbalance motivates the design of our learning objectives (Sec.~\ref{sec:pro-loss}).

\section{Methodology}
\subsection{Problem Formulation}

We formulate Omni-Human Pose Estimation (Omni-HPE) as a joint learning problem of coordinate regression and semantic attribution. Given an input image $I \in \mathbb{R}^{H \times W \times 3}$, our objective is to predict a set $K$ keypoints $\mathcal{P} = \{\mathbf{p}_i\}_{i=1}^K$, their corresponding confidence $s_i$ and semantics $c_i$:
\begin{equation}
    \mathbf{p}_i = (\mathbf{x}_i, s_i, c_i),
\end{equation}
where $\mathbf{x}_i \in \mathbb{R}^2$ denotes the spatial coordinates, $s_i \in [0, 1]$ represents the prediction confidence, and $c_i \in \mathcal{C}$ represents the semantic class and the label space $\mathcal{C} = \{\texttt{Bio}, \texttt{Pros}, \texttt{Abs}\}$, corresponding to {Biological}, {Prosthetic}, and {Absent} states, respectively. Note that $\texttt{Bio}$ includes both intact and residual limbs. They can be easily distinguished by their keypoint ordering.

For keypoints labeled as \texttt{Abs}, the spatial coordinate does not correspond to a physically existing anatomical or mechanical point. Therefore, coordinate supervision for these keypoints is masked out during training, while their semantic type labels are retained for classification.

\subsection{Proposed Learning Objective: ProLoss}
\label{sec:pro-loss}
Learning to estimate poses for individuals with limb differences presents two primary challenges. First, the severe class imbalance between biological and prosthetic or absent joints causes standard loss formulations to heavily bias the model toward the majority class. To address this distribution bias, we introduce a re-weighting strategy for semantic categorization. Second, standard objective functions assume an independent and identically distributed (i.i.d.) topology among keypoints. This assumption might lead to hallucinated keypoints. To jointly resolve these issues, we propose \textbf{ProLoss}, a structure-aware objective function composed of three terms: Location Regression ($\mathcal{L}_{reg}$) and Semantic Categorization ($\mathcal{L}_{cls}$), alongside a Bio-Contrastive Boundary Loss ($\mathcal{L}_{bio}$).

\subsubsection{Distribution-Aligned Supervision:}
\label{sec:dist_aligned}

Inspired by the strategies that address class imbalance, such as the Class-Balanced (CB) loss \cite{cui2019class} and Focal Loss \cite{Lin_2017_ICCV}, we employ a decoupled, inverse-square-root weighting strategy to effectively promote the influence of rare instances.

\paragraph{Semantic Categorization Weighting.}
For the semantic classification of each keypoint $i$, we apply a point-wise normalized weight to prevent the majority biological class from dominating the cross-entropy loss. The classification weight $\omega_{i, c}^{cls}$ for class $c$ is inversely proportional to the square root of its frequency:
\begin{equation}
    \omega_{i, c}^{cls} = \frac{1 / \sqrt{N_{i, c}}}{\frac{1}{|\mathcal{C}_{valid}|} \sum_{j \in \mathcal{C}_{valid}} (1 / \sqrt{N_{i, j}})},
\end{equation}
where $N_{i, c}$ is the number of training samples for class $c$ at keypoint $i$, and $\mathcal{C}_{valid}$ represents the valid classes present for that specific joint.

With the semantic classification weight $\omega_{i, c}^{cls}$, we formulate our combined objective. Since spatial coordinate regression is unaffected by the semantic long-tail distribution, we do not apply additional re-weighting to the regression branch. The localization loss $\mathcal{L}_{reg}$ and the distribution-aligned semantic loss $\mathcal{L}_{cls}$ are formally defined as:
\begin{equation}
    \mathcal{L}_{reg} = \frac{1}{K} \sum_{i=1}^{K} \mathbbm{1}(v_i > 0 \text{ \& } c_i^* \neq \text{absent}) \cdot \ell(\mathbf{p}_i, \hat{\mathbf{p}}_i),
\end{equation}
\begin{equation}
    \mathcal{L}_{cls} = - \frac{1}{K} \sum_{i=1}^{K} \mathbbm{1}(v_i > 0) \cdot \omega_{i, c_i^*}^{cls} \sum_{c \in \mathcal{C}} y_{i,c}^* \log(\hat{y}_{i,c}),
\end{equation}
where $\mathbbm{1}(\cdot)$ denotes the indicator function, equal to $1$ if the specified condition is true and $0$ otherwise. $v_i > 0$ denotes visibility, $c_i^*$ is the ground-truth semantic class, and $\ell(\cdot, \cdot)$ represents the spatial regression loss (\eg, MSE).

\subsubsection{Bio-Contrastive Boundary Loss ($\mathcal{L}_{bio}$):}
\label{sec:bio_contrastive}

Limb deficiencies impose strict anatomical mutual exclusivity: 
Each limb branch in the kinematic structure admits only one anatomically valid terminal state. Thus, any residual points or additional joint predictions that violate biological constraints are physically infeasible.
To enforce this, we directly leverage intra-limb semantic mutual exclusivity. For an annotated residual keypoint $r \in \mathcal{R}$, we define its \textit{Mutually Exclusive Set} $\Omega_r$ as the set of all keypoints that are biologically incompatible with $r$ in the same limb branch:
\begin{equation}
    \Omega_r = \{ j \in \mathcal{B}_{L(r)} \mid j \neq r \text{ \& } j \bot r \},
\end{equation}
where $\mathcal{B}_{L(r)}$ denotes all keypoints belonging to limb branch $L(r)$, and $\bot$ indicates incompatibility. 
Our Bio-Contrastive Loss maximizes the probability of the valid residual keypoint $r$ with respect to its mutually exclusive set. For a residual keypoint $r$, we define its normalized biological confidence $\mathcal{P}_r$ as:
\begin{equation}
    \mathcal{P}_r = \frac{\exp(p_r^{\text{bio}} / \tau)}{\exp(p_r^{\text{bio}} / \tau) + \sum_{j \in \Omega_r} \mathbbm{1}(v_j > 0) \cdot \exp(p_j^{\text{bio}} / \tau)},
\end{equation}
where $\tau$ is a temperature hyperparameter. The resulting loss $\mathcal{L}_{bio}$ is then formulated as the negative log-likelihood over all the annotated residual keypoints:
\begin{equation}
    \mathcal{L}_{bio} = - \frac{1}{|\mathcal{R}|} \sum_{r \in \mathcal{R}} \mathbbm{1}(v_r > 0 \text{ \& } c_r^* = \textit{Biological}) \cdot \log \left( \mathcal{P}_r \right),
\end{equation}
where $c_r^* = \texttt{biological}$ ensures that the contrastive objective only operates on keypoints representing the true biological end of the residual limb. 
Minimizing loss $\mathcal{L}_{bio}$ encourages the model to increase the probability of residual limb types while suppressing intact limb predictions for their descendant joints.
As a result, this loss efficiently suppresses anatomical hallucinations and improves the classification recall for both prosthetic (\texttt{Pros}) and absent (\texttt{Abs}) categories.

Overall, the total objective is formulated as:
\begin{equation}
    \mathcal{L}_{total} = \lambda_{reg}\mathcal{L}_{reg} + \lambda_{cls}\mathcal{L}_{cls} + \lambda_{bio}\mathcal{L}_{bio}.
\end{equation}

\section{Experiments}
\label{sec:experiments}

\begin{table*}[t]
\centering
\caption{Performance comparison on the ProPose Test Set. The metrics are decoupled into geometric precision (AP and AR across multiple thresholds) and semantic diagnostics (Type-Acc for Total, Bio, Pros, and Abs). The Abs accuracy is computed only over non-residual keypoints, excluding Absent labels on residual keypoints, to avoid the large number of structurally absent residual labels dominating the metric.}
\label{tab:propose_benchmark}

\setlength{\tabcolsep}{1.2pt}
\resizebox{\textwidth}{!}{
\scriptsize
\begin{tabular}{c c c c | cccccc cccc}
\toprule
\multirow{5}{*}{\textbf{Method}} & \multirow{5}{*}{\textbf{Training}} & \multirow{5}{*}{\textbf{Backbone}} & \multirow{5}{*}{\textbf{Input Size}} & \multicolumn{10}{c}{\textbf{ProPose Test Set}} \\
\cmidrule{5-14}
& & & & \multicolumn{6}{c|}{\textbf{Geometric (AP \& AR)}} & \multicolumn{4}{c}{\begin{tabular}{@{}c@{}}\textbf{Semantic} \\ \textbf{(Type-Acc)}\end{tabular}} \\
\cmidrule{5-14}
& & & & AP & AP$^{50}$ & AP$^{75}$ & AR & AR$^{50}$ & AR$^{75}$ & Total & Bio & Pros & Abs \\
\midrule

\multirow{8}{*}{YoloxPose} 
 & \cellcolor{rowours} & \cellcolor{rowours}YoloxPose-T & \cellcolor{rowours}416$\times$416 & \cellcolor{rowours}41.5 & \cellcolor{rowours}60.3 & \cellcolor{rowours}46.2 & \cellcolor{rowours}46.9 & \cellcolor{rowours}65.3 & \cellcolor{rowours}51.5 & \cellcolor{rowours} $-$ & \cellcolor{rowours}$-$ & \cellcolor{rowours}$-$ & \cellcolor{rowours}$-$ \\
 & \cellcolor{rowours} & \cellcolor{rowours}YoloxPose-S & \cellcolor{rowours}640$\times$640 & \cellcolor{rowours}48.8 & \cellcolor{rowours}66.2 & \cellcolor{rowours}54.8 & \cellcolor{rowours}55.1 & \cellcolor{rowours}71.3 & \cellcolor{rowours}60.6 & \cellcolor{rowours} $-$ & \cellcolor{rowours} $-$ & \cellcolor{rowours} $-$ & \cellcolor{rowours} $-$ \\
 & \cellcolor{rowours} & \cellcolor{rowours}YoloxPose-M & \cellcolor{rowours}640$\times$640 & \cellcolor{rowours}52.2 & \cellcolor{rowours}68.7 & \cellcolor{rowours}59.1 & \cellcolor{rowours}58.5 & \cellcolor{rowours}73.3 & \cellcolor{rowours}64.8 & \cellcolor{rowours}$-$ & \cellcolor{rowours}$-$ & \cellcolor{rowours}$-$ & \cellcolor{rowours}$-$ \\
 & \cellcolor{rowours}\multirow{-4}{*}{Pretrain} & \cellcolor{rowours}YoloxPose-L & \cellcolor{rowours}640$\times$640 & \cellcolor{rowours}53.4 & \cellcolor{rowours}69.3 & \cellcolor{rowours}60.2 & \cellcolor{rowours}59.8 & \cellcolor{rowours}73.8 & \cellcolor{rowours}65.8 & \cellcolor{rowours}$-$ & \cellcolor{rowours}$-$ & \cellcolor{rowours}$-$ & \cellcolor{rowours}$-$ \\
\cmidrule{2-14} 
 & & YoloxPose-T & 416$\times$416 & 41.9 & 67.3 & 43.0 & 49.0 & 72.5 & 50.4 & 82.9 & 89.7 & 53.8 & 27.7 \\
 & & YoloxPose-S & 640$\times$640 & 50.5 & 74.3 & 53.6 & 58.6 & 79.3 & 61.4 & 86.8 & 90.4 & 64.6 & 52.1 \\
 & & YoloxPose-M & 640$\times$640 & 52.2 & 76.3 & 55.6 & 60.5 & 82.0 & 63.4 & 88.5 & 92.3 & 66.3 & 54.0 \\
 & \multirow{-4}{*}{$+$ProLoss} & YoloxPose-L & 640$\times$640 & 53.3 & 76.7 & 56.3 & 61.0 & 81.2 & 63.5 & 89.0 & 92.6 & 69.4 & 56.6 \\
\midrule

\multirow{10}{*}{\begin{tabular}{@{}c@{}}Swin \\ Transformer\end{tabular}} 
 & \cellcolor{rowours} & \cellcolor{rowours}Swin-T & \cellcolor{rowours}256$\times$192 & \cellcolor{rowours}74.7 & \cellcolor{rowours}95.7 & \cellcolor{rowours}83.5 & \cellcolor{rowours}78.9 & \cellcolor{rowours}96.1 & \cellcolor{rowours}85.9 & \cellcolor{rowours}$-$ & \cellcolor{rowours}$-$ & \cellcolor{rowours}$-$ & \cellcolor{rowours}$-$ \\
 & \cellcolor{rowours} & \cellcolor{rowours}Swin-B & \cellcolor{rowours}256$\times$192 & \cellcolor{rowours}76.4 & \cellcolor{rowours}95.8 & \cellcolor{rowours}85.6 & \cellcolor{rowours}80.5 & \cellcolor{rowours}96.1 & \cellcolor{rowours}87.2 & \cellcolor{rowours}$-$ & \cellcolor{rowours}$-$ & \cellcolor{rowours}$-$ & \cellcolor{rowours}$-$ \\
 & \cellcolor{rowours} & \cellcolor{rowours}Swin-B & \cellcolor{rowours}384$\times$288 & \cellcolor{rowours}76.1 & \cellcolor{rowours}95.7 & \cellcolor{rowours}85.6 & \cellcolor{rowours}80.5 & \cellcolor{rowours}96.3 & \cellcolor{rowours}87.4 & \cellcolor{rowours}$-$ & \cellcolor{rowours}$-$ & \cellcolor{rowours}$-$ & \cellcolor{rowours}$-$ \\
 & \cellcolor{rowours} & \cellcolor{rowours}Swin-L & \cellcolor{rowours}256$\times$192 & \cellcolor{rowours}76.1 & \cellcolor{rowours}95.8 & \cellcolor{rowours}85.7 & \cellcolor{rowours}80.5 & \cellcolor{rowours}96.5 & \cellcolor{rowours}87.2 & \cellcolor{rowours}$-$ & \cellcolor{rowours}$-$ & \cellcolor{rowours}$-$ & \cellcolor{rowours}$-$ \\
 & \cellcolor{rowours}\multirow{-5}{*}{Pretrain} & \cellcolor{rowours}Swin-L & \cellcolor{rowours}384$\times$288 & \cellcolor{rowours}76.8 & \cellcolor{rowours}95.8 & \cellcolor{rowours}86.6 & \cellcolor{rowours}81.3 & \cellcolor{rowours}96.4 & \cellcolor{rowours}88.4 & \cellcolor{rowours}$-$ & \cellcolor{rowours}$-$ & \cellcolor{rowours}$-$ & \cellcolor{rowours}$-$ \\
\cmidrule{2-14} 
 & & Swin-T & 256$\times$192 & 85.7 & 96.5 & 91.9 & 88.1 & 97.7 & 93.2 & 91.5 & 94.0 & 71.9 & 68.0 \\
 & & Swin-B & 256$\times$192 & 87.7 & 96.7 & 93.3 & 89.8 & 97.7 & 94.2 & 96.9 & 98.7 & 86.2 & 82.9 \\
 & & Swin-B & 384$\times$288 & 88.7 & 96.7 & 93.4 & 90.4 & 97.7 & 94.3 & \textbf{97.2} & \textbf{99.0} & 86.0 & 82.6 \\
 & & Swin-L & 256$\times$192 & 86.4 & 96.7 & 92.2 & 88.8 & 97.5 & 93.7 & 96.7 & 98.6 & 83.6 & 82.7 \\
 & \multirow{-5}{*}{$+$ProLoss} & Swin-L & 384$\times$288 & 87.1 & 96.6 & 92.3 & 89.0 & 97.5 & 93.4 & 97.1 & 98.8 & 85.0 & 83.3 \\
\midrule

\multirow{8}{*}{RTMPose} 
 & \cellcolor{rowours} & \cellcolor{rowours}RTMPose-T & \cellcolor{rowours}256$\times$192 & \cellcolor{rowours}70.7 & \cellcolor{rowours}92.7 & \cellcolor{rowours}80.8 & \cellcolor{rowours}75.5 & \cellcolor{rowours}94.0 & \cellcolor{rowours}83.4 & \cellcolor{rowours}$-$ & \cellcolor{rowours}$-$ & \cellcolor{rowours}$-$ & \cellcolor{rowours}$-$ \\
 & \cellcolor{rowours} & \cellcolor{rowours}RTMPose-S & \cellcolor{rowours}256$\times$192 & \cellcolor{rowours}73.3 & \cellcolor{rowours}94.7 & \cellcolor{rowours}83.2 & \cellcolor{rowours}78.1 & \cellcolor{rowours}95.5 & \cellcolor{rowours}85.5 & \cellcolor{rowours}$-$ & \cellcolor{rowours}$-$ & \cellcolor{rowours}$-$ & \cellcolor{rowours}$-$ \\
 & \cellcolor{rowours} & \cellcolor{rowours}RTMPose-M & \cellcolor{rowours}256$\times$192 & \cellcolor{rowours}76.1 & \cellcolor{rowours}95.8 & \cellcolor{rowours}85.6 & \cellcolor{rowours}80.8 & \cellcolor{rowours}96.3 & \cellcolor{rowours}87.8 & \cellcolor{rowours}$-$ & \cellcolor{rowours}$-$ & \cellcolor{rowours}$-$ & \cellcolor{rowours}$-$ \\
 & \cellcolor{rowours}\multirow{-4}{*}{Pretrain} & \cellcolor{rowours}RTMPose-L & \cellcolor{rowours}256$\times$192 & \cellcolor{rowours}77.2 & \cellcolor{rowours}95.9 & \cellcolor{rowours}86.8 & \cellcolor{rowours}81.9 & \cellcolor{rowours}96.8 & \cellcolor{rowours}88.9 & \cellcolor{rowours}$-$ & \cellcolor{rowours}$-$ & \cellcolor{rowours}$-$ & \cellcolor{rowours}$-$ \\
\cmidrule{2-14} 
 & & RTMPose-T & 256$\times$192 & 83.0 & 95.7 & 89.2 & 85.1 & 96.8 & 90.4 & 93.8 & 96.3 & 75.3 & 72.4 \\
 & & RTMPose-S & 256$\times$192 & 86.1 & 96.6 & 92.3 & 87.9 & 97.4 & 93.0 & 94.7 & 96.8 & 78.5 & 78.0 \\
 & & RTMPose-M & 256$\times$192 & 88.8 & \textbf{96.8} & 93.6 & 90.4 & 97.8 & 94.8 & 95.8 & 97.8 & 82.5 & 80.3 \\
 & \multirow{-4}{*}{$+$ProLoss} & RTMPose-L & 256$\times$192 & 89.9 & \textbf{96.8} & \textbf{94.6} & 91.3 & 97.9 & 95.1 & 96.5 & 98.2 & 85.3 & 83.7 \\
\midrule

\multirow{6}{*}{ViTPose} 
 & \cellcolor{rowours} & \cellcolor{rowours}ViT-S & \cellcolor{rowours}256$\times$192 & \cellcolor{rowours}76.3 & \cellcolor{rowours}95.7 & \cellcolor{rowours}85.6 & \cellcolor{rowours}80.6 & \cellcolor{rowours}96.1 & \cellcolor{rowours}87.3 & \cellcolor{rowours}$-$ & \cellcolor{rowours}$-$ & \cellcolor{rowours}$-$ & \cellcolor{rowours}$-$ \\
 & \cellcolor{rowours} & \cellcolor{rowours}ViT-B & \cellcolor{rowours}256$\times$192 & \cellcolor{rowours}78.1 & \cellcolor{rowours}95.7 & \cellcolor{rowours}87.7 & \cellcolor{rowours}82.6 & \cellcolor{rowours}96.8 & \cellcolor{rowours}89.7 & \cellcolor{rowours}$-$ & \cellcolor{rowours}$-$ & \cellcolor{rowours}$-$ & \cellcolor{rowours}$-$ \\
 & \cellcolor{rowours}\multirow{-3}{*}{Pretrain} & \cellcolor{rowours}ViT-L & \cellcolor{rowours}256$\times$192 & \cellcolor{rowours}80.5 & \cellcolor{rowours}96.9 & \cellcolor{rowours}90.8 & \cellcolor{rowours}85.1 & \cellcolor{rowours}97.6 & \cellcolor{rowours}82.3 & \cellcolor{rowours}$-$ & \cellcolor{rowours}$-$ & \cellcolor{rowours}$-$ & \cellcolor{rowours}$-$ \\
\cmidrule{2-14} 
 & & ViT-S & 256$\times$192 & 81.7 & 95.5 & 88.0 & 84.5 & 96.5 & 89.9 & 93.7 & 95.9 & 75.5 & 73.0 \\
 & & ViT-B & 256$\times$192 & 82.0 & 95.5 & 88.9 & 84.7 & 96.8 & 90.2 & 95.0 & 97.5 & 75.5 & 75.8 \\
 & \multirow{-3}{*}{$+$ProLoss} & ViT-L & 256$\times$192 & \textbf{90.0} & 96.7 & 94.5 & \textbf{91.9} & \textbf{98.0} & \textbf{95.7} & 97.0 & 98.2 & \textbf{92.3} & \textbf{85.9} \\
\bottomrule
\end{tabular}
}
\end{table*}

\subsection{Implementation Details}
\label{sec:impl_details}

We implement all baseline methods and our proposed framework using the MMPose codebase \cite{mmpose2020}. To ensure a fair and rigorous evaluation, identical training and testing splits are used across all methods. Since Omni-Pose involves both spatial localization and semantic classification, we evaluate models on both aspects. For geometric accuracy, we report the standard Mean Average Precision (mAP). Notably, \texttt{Absent} keypoints are excluded from mAP calculations (despite inferred visibility $v=2$) since their locations are physically meaningless. Conversely, semantic accuracy evaluates attribute identification across all predicted types, including \texttt{Absent}.

\paragraph{Hyperparameter Settings.}
For all experiments using ProLoss, we set $\lambda_{reg}=1$, $\lambda_{cls}=0.05$, $\lambda_{bio}=0.01$, and $\tau=0.2$. The classification-related losses are weighted to remain approximately one order of magnitude smaller than the coordinate regression loss. Empirically, we observe that AP and Type-Acc remain stable under moderate variations of these hyperparameters, indicating that ProLoss does not require task-specific tuning.
\subsection{Benchmark Results}
\label{sec:benchmark}
\begin{figure*}[t!]
  \centering
  \includegraphics[width=\linewidth]{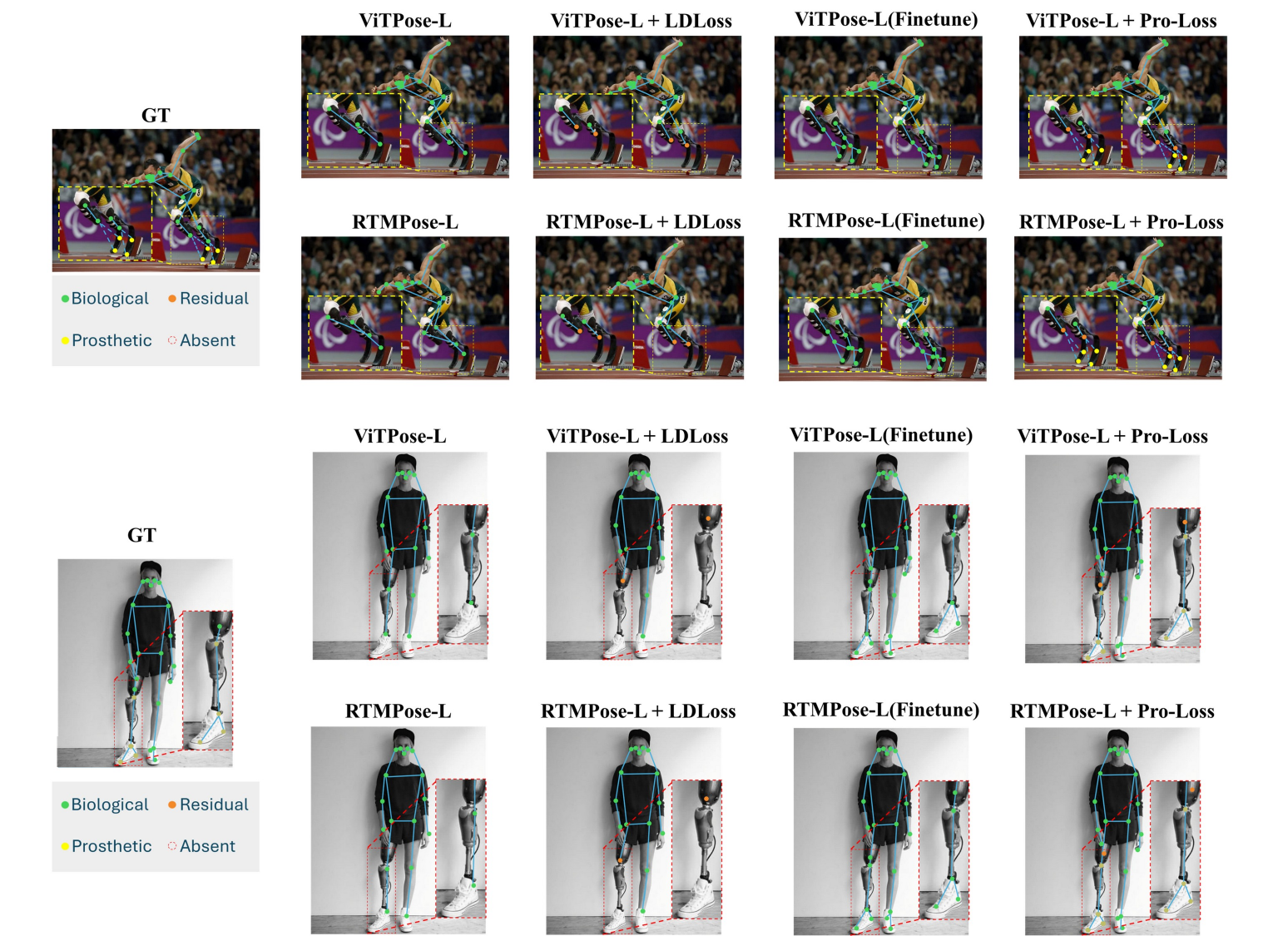}
    \caption{Qualitative comparison between baseline methods and our ProLoss. Pre-trained baselines hallucinate biological limbs, while LDLoss completely ignores prostheses. Standard fine-tuning on keypoint coordinates fails to differentiate semantic types, leading to hallucinated absent joints. In contrast, our method accurately categorizes prosthetic and absent points to successfully reconstruct the kinematic structure.}
  \label{fig:qualitative_comparison}
\end{figure*}

We establish a comprehensive baseline on our ProPose dataset by evaluating state-of-the-art 2D pose estimators \cite{xu2022vitpose, jiang2023rtmpose, maji2022yolo, xiong2022swin}, specifically comparing off-the-shelf pre-trained models against those fine-tuned with our ProLoss. We also evaluate fine-tuning with a standard spatial regression loss and LDLoss (visually compared in Fig.~\ref{fig:qualitative_comparison}). As shown in Table \ref{tab:propose_benchmark}, models fine-tuned with ProLoss significantly outperform their pre-trained counterparts in spatial metrics. Additionally, top-down architectures demonstrate robust geometric precision, maintaining high mAP despite the increased complexity of localizing 6 extra extremity keypoints and predicting semantic types. Notably, ViTPose-L achieves a peak mAP of \textbf{90.0}. However, standard models struggle to accurately classify minority joint states. While ViTPose-L performs best (\textbf{92.3\%} for prosthetic and \textbf{85.9\%} for absent keypoints), others experience sharp drops, highlighting the inherent difficulty of distinguishing specialized mechanical structures from biological limbs. 

To evaluate cross-dataset generalization, we manually annotated an independent test set from ProGait \cite{yin2025progait} videos. To ensure postural diversity, we sampled and manually filtered 2-3 frames from the initial and final quarters of each clip. As shown in Table \ref{tab:progait_benchmark}, models maintain stable or higher metrics than on ProPose. Crucially, ProLoss-finetuned models consistently outperform pre-trained baselines. ViTPose-L remains the most robust against long-tail distributions, achieving \textbf{90.2\%} prosthetic accuracy, despite Swin-B marginally surpassing it on \texttt{Bio} joints. Ultimately, this confirms that ProPose-trained models readily generalize to novel real-world limb differences.

\begin{figure*}[!t]
    \centering
    \includegraphics[width=\textwidth,height=0.82\textheight,keepaspectratio]{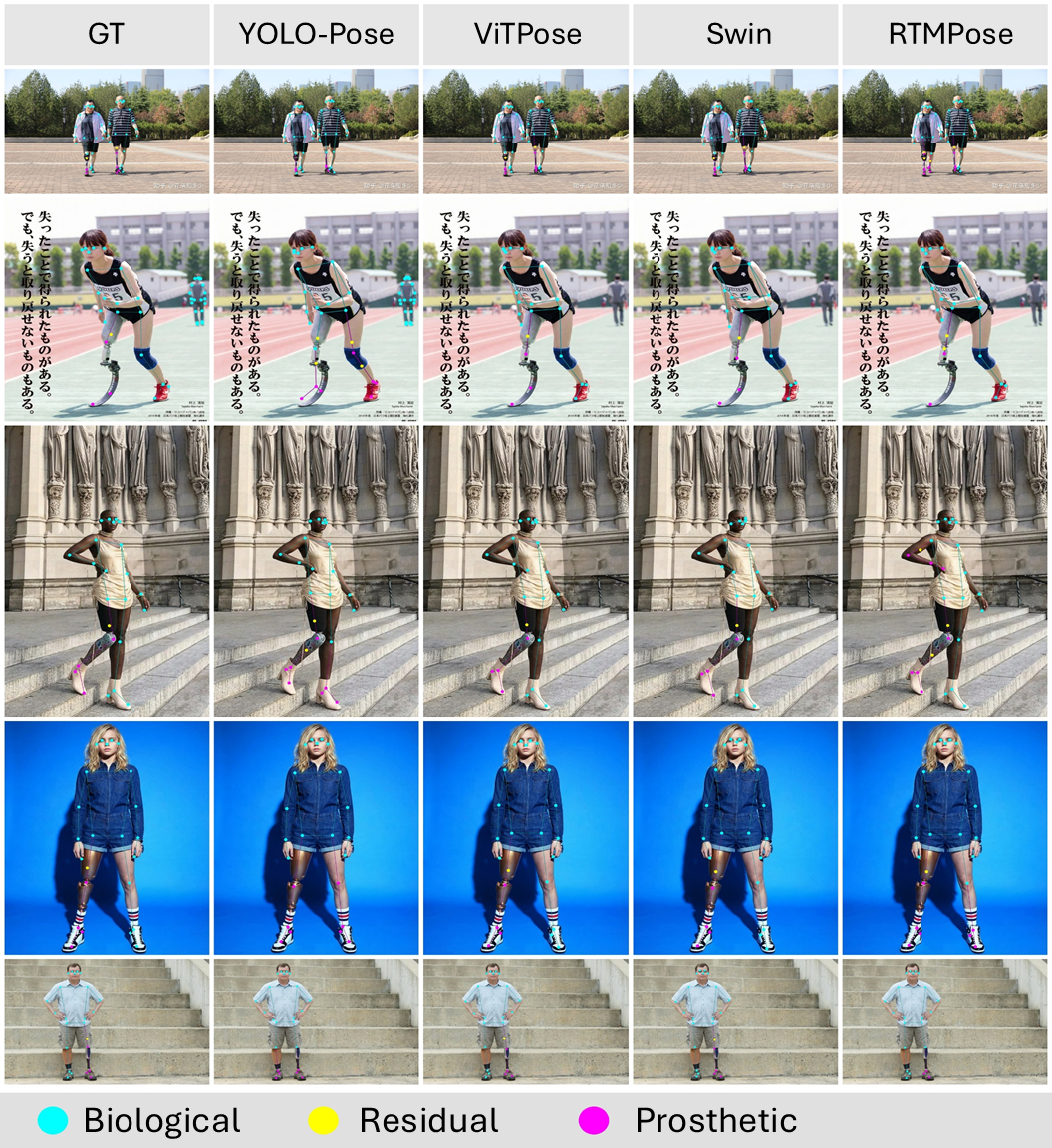}
    \caption{Qualitative results of applying our proposed loss function to different pose estimation models. From left to right: ground truth, YOLO-Pose, ViTPose, Swin-Pose, and RTMPose trained with our loss on the ProPose dataset.}
    \label{fig:dataset_overview}
\end{figure*}

\begin{table*}[t]
\centering
\caption{Benchmarking results on ProGait for cross-dataset evaluation.}
\label{tab:progait_benchmark}
\scriptsize
\setlength{\tabcolsep}{1.5pt}
\resizebox{\textwidth}{!}{
\begin{tabular}{c c c c | cccccc cccc}
\toprule
\multirow{5}{*}{\textbf{Method}} & \multirow{5}{*}{\textbf{Training}} & \multirow{5}{*}{\textbf{Backbone}} & \multirow{5}{*}{\textbf{Input Size}} & \multicolumn{10}{c}{\textbf{ProGait}} \\
\cmidrule{5-14}
& & & & \multicolumn{6}{c|}{\textbf{Geometric (AP \& AR)}} & \multicolumn{4}{c}{\begin{tabular}{@{}c@{}}\textbf{Semantic} \\ \textbf{(Type-Acc)}\end{tabular}} \\
\cmidrule{5-14}
& & & & AP & AP$^{50}$ & AP$^{75}$ & AR & AR$^{50}$ & AR$^{75}$ & Total & Bio & Pros & Abs \\
\midrule

\multirow{4}{*}{YoloxPose} 
 & \cellcolor{rowours} & \cellcolor{rowours}YoloxPose-T & \cellcolor{rowours}416$\times$416 & \cellcolor{rowours}44.2 & \cellcolor{rowours}58.7 & \cellcolor{rowours}55.5 & \cellcolor{rowours}75.5 & \cellcolor{rowours}95.7 & \cellcolor{rowours}90.6 & \cellcolor{rowours}$-$ & \cellcolor{rowours}$-$ & \cellcolor{rowours}$-$ & \cellcolor{rowours}$-$ \\
 & \cellcolor{rowours}\multirow{-2}{*}{Pretrain} & \cellcolor{rowours}YoloxPose-L & \cellcolor{rowours}640$\times$640 & \cellcolor{rowours}42.9 & \cellcolor{rowours}52.4 & \cellcolor{rowours}51.7 & \cellcolor{rowours}84.1 & \cellcolor{rowours}98.0 & \cellcolor{rowours}96.8 & \cellcolor{rowours}$-$ & \cellcolor{rowours}$-$ & \cellcolor{rowours}$-$ & \cellcolor{rowours}$-$ \\
\cmidrule{2-14}
 & & YoloxPose-T & 416$\times$416 & 48.6 & 71.0 & 54.5 & 65.9 & 91.2 & 74.5 & 62.5 & 75.4 & 44.4 & 95.7 \\
 & \multirow{-2}{*}{$+$ProLoss} & YoloxPose-L & 640$\times$640 & 69.8 & 89.4 & 83.9 & 78.1 & 96.8 & 91.1 & 73.6 & 87.7 & 54.9 & 96.7 \\
\midrule

\multirow{4}{*}{\begin{tabular}{@{}c@{}}Swin \\ Transformer\end{tabular}} 
 & \cellcolor{rowours} & \cellcolor{rowours}Swin-T & \cellcolor{rowours}256$\times$192 & \cellcolor{rowours}75.3 & \cellcolor{rowours}97.9 & \cellcolor{rowours}93.3 & \cellcolor{rowours}79.6 & \cellcolor{rowours}98.4 & \cellcolor{rowours}94.3 & \cellcolor{rowours}$-$ & \cellcolor{rowours}$-$ & \cellcolor{rowours}$-$ & \cellcolor{rowours}$-$ \\
 & \cellcolor{rowours}\multirow{-2}{*}{Pretrain} & \cellcolor{rowours}Swin-L & \cellcolor{rowours}384$\times$288 & \cellcolor{rowours}73.7 & \cellcolor{rowours}96.8 & \cellcolor{rowours}90.5 & \cellcolor{rowours}78.1 & \cellcolor{rowours}97.2 & \cellcolor{rowours}92.7 & \cellcolor{rowours}$-$ & \cellcolor{rowours}$-$ & \cellcolor{rowours}$-$ & \cellcolor{rowours}$-$ \\
\cmidrule{2-14}
 & & Swin-T & 256$\times$192 & 87.5 & 98.9 & 96.8 & 90.3 & 99.4 & 97.6 & 61.9 & 63.5 & 59.0 & 96.8 \\
 & \multirow{-2}{*}{$+$ProLoss} & Swin-L & 384$\times$288 & 88.5 & 98.9 & 97.8 & 91.1 & 99.4 & 98.5 & 92.3 & 97.9 & 84.7 & 97.9 \\
\midrule

\multirow{4}{*}{RTMPose} 
 & \cellcolor{rowours} & \cellcolor{rowours}RTMPose-T & \cellcolor{rowours}256$\times$192 & \cellcolor{rowours}75.7 & \cellcolor{rowours}97.7 & \cellcolor{rowours}93.3 & \cellcolor{rowours}80.7 & \cellcolor{rowours}98.9 & \cellcolor{rowours}95.0 & \cellcolor{rowours}$-$ & \cellcolor{rowours}$-$ & \cellcolor{rowours}$-$ & \cellcolor{rowours}$-$ \\
 & \cellcolor{rowours}\multirow{-2}{*}{Pretrain} & \cellcolor{rowours}RTMPose-L & \cellcolor{rowours}256$\times$192 & \cellcolor{rowours}79.7 & \cellcolor{rowours}96.8 & \cellcolor{rowours}95.6 & \cellcolor{rowours}83.9 & \cellcolor{rowours}98.0 & \cellcolor{rowours}96.6 & \cellcolor{rowours}$-$ & \cellcolor{rowours}$-$ & \cellcolor{rowours}$-$ & \cellcolor{rowours}$-$ \\
\cmidrule{2-14}
 & & RTMPose-T & 256$\times$192 & 84.9 & 97.9 & 95.8 & 87.3 & 99.0 & 96.7 & 85.0 & 94.8 & 72.5 & 99.1 \\
 & \multirow{-2}{*}{$+$ProLoss} & RTMPose-L & 256$\times$192 & 89.5 & \textbf{99.0} & \textbf{97.9} & 91.6 & 99.3 & \textbf{98.7} & 90.3 & 97.2 & 80.6 & 99.1 \\
\midrule

\multirow{4}{*}{ViTPose} 
 & \cellcolor{rowours} & \cellcolor{rowours}ViT-S & \cellcolor{rowours}256$\times$192 & \cellcolor{rowours}75.2 & \cellcolor{rowours}96.5 & \cellcolor{rowours}92.2 & \cellcolor{rowours}76.9 & \cellcolor{rowours}97.1 & \cellcolor{rowours}93.5 & \cellcolor{rowours}$-$ & \cellcolor{rowours}$-$ & \cellcolor{rowours}$-$ & \cellcolor{rowours}$-$ \\
 & \cellcolor{rowours}\multirow{-2}{*}{Pretrain} & \cellcolor{rowours}ViT-L & \cellcolor{rowours}256$\times$192 & \cellcolor{rowours}76.8 & \cellcolor{rowours}96.7 & \cellcolor{rowours}94.2 & \cellcolor{rowours}81.8 & \cellcolor{rowours}97.5 & \cellcolor{rowours}95.6 & \cellcolor{rowours}$-$ & \cellcolor{rowours}$-$ & \cellcolor{rowours}$-$ & \cellcolor{rowours}$-$ \\
\cmidrule{2-14}
 & & ViT-S & 256$\times$192 & 82.5 & 97.6 & 93.2 & 85.1 & 98.2 & 94.1 & 87.8 & 94.0 & 79.4 & 97.1 \\
 & \multirow{-2}{*}{$+$ProLoss} & ViT-L & 256$\times$192 & \textbf{90.9} & \textbf{99.0} & \textbf{97.9} & \textbf{92.9} & 99.4 & \textbf{98.7} & \textbf{94.4} & 97.1 & \textbf{90.2} & 98.3 \\
\bottomrule
\end{tabular}
}
\end{table*}

\begin{table*}[t]
\centering
\caption{Ablation study of model architecture and ProLoss components. We evaluate the impact of gradient flow (Detach), Semantic Categorization Weighting (SCW), and Bio-Contrastive Boundary Loss ($\mathcal{L}_{bio}$). For $\mathcal{L}_{bio}$, we compare joint training versus stage-wise finetuning (staged). Res-Bio explicitly evaluates the classification accuracy on the specific residual endpoint keypoints.}
\label{tab:comprehensive_ablation}
\scriptsize
\setlength{\tabcolsep}{2pt}
\begin{tabular}{c|cc|c||cccc||cccc|c}
\toprule
\multirow{2}{*}{\textbf{Detach}} & \multicolumn{2}{c|}{\textbf{SCW}} & \multirow{2}{*}{\textbf{$\mathcal{L}_{bio}$}} & \multicolumn{4}{c||}{\textbf{Geometric Metrics}} & \multicolumn{4}{c|}{\textbf{Type-Acc (\%)}} & \textbf{Kp Acc} \\
 & \textbf{Cls} & \textbf{Reg} &  & \textbf{AP} & \textbf{$\text{AP}_{50}$} & \textbf{$\text{AP}_{75}$} & \textbf{AR} & \textbf{Total} & \textbf{Bio} & \textbf{Pros} & \textbf{Abs} & \textbf{$\textbf{Type}_{Res}$=Bio} \\
\midrule
\checkmark & -- & -- & -- & 89.7 & 96.6 & 94.4 & 91.9 & 96.6 & 98.3 & 85.7 & 82.4 & 76.6 \\
-- & -- & -- & -- & 89.9 & \textbf{96.7} & 94.4 & \textbf{92.0} & \textbf{97.4} & \textbf{99.1} & 89.7 & 81.9 & 78.5 \\
\midrule
-- & \checkmark & -- & -- & \textbf{90.0} & \textbf{96.7} & \textbf{94.5} & \textbf{92.0} & 97.3 & 98.7 & 90.7 & 84.6 & 81.1 \\
-- & \checkmark & \checkmark & -- & 89.7 & \textbf{96.7} & 94.4 & 91.8 & 97.3 & 98.6 & 91.5 & 85.1 & 81.7 \\
\midrule
-- & \checkmark & -- & Joint & 89.6 & \textbf{96.7} & 94.4 & 91.6 & 97.1 & 98.4 & 91.5 & 84.9 & 81.5 \\
-- & \checkmark & -- & Staged & \textbf{90.0} & \textbf{96.7} & \textbf{94.5} & 91.9 & 97.0 & 98.3 & \textbf{91.7} & \textbf{85.6} & \textbf{86.9} \\
\bottomrule
\end{tabular}
\end{table*}

\subsection{Ablation Study}
\label{sec:ablation}

To validate our architecture and ProLoss components, we perform an ablation study using ViTPose-L on ProPose. As Table~\ref{tab:comprehensive_ablation} shows, results are decoupled into geometric precision and semantic diagnostics. Notably, we include the $\text{Acc}_{\texttt{Res-Bio}}$ metric to explicitly track the detection of physical residual stumps, preventing this rare subset from being overshadowed by dominant biological joints.

\paragraph{Impact of Joint Feature Learning (Detach vs. Attached).}
We first investigate the interaction between spatial coordinate regression and semantic type classification. Specifically, we compare a detached classification head, where gradients do not flow back to the shared backbone, against an attached, jointly trained configuration. As shown in Table \ref{tab:comprehensive_ablation}, when the classification head is detached, the model achieves an AP of 89.7\% and a \texttt{Pros} accuracy of 85.7\%. Conversely, allowing gradients from the classification loss to update the shared backbone yields simultaneous improvements in both metrics. Notably, the overall AP increases by 0.2\%, alongside a substantial 4.2\% surge in \texttt{Pros} accuracy. This firmly indicates that semantic attribute prediction acts as a beneficial auxiliary task, actively enhancing spatial feature representations rather than degrading geometric localization performance. Furthermore, it demonstrates that the standard feature embeddings extracted by the backbone inherently lack sufficient semantic information for accurate type classification without task-specific gradient feedback.

\paragraph{Effectiveness of Semantic Categorization Weighting.}
Compared to the standard attached baseline without re-weighting, applying Semantic Categorization Weighting exclusively to the classification head significantly improves the semantic accuracy of minority classes (\texttt{Pros} +1.0\%, \texttt{Abs} +2.7\%, \texttt{Res-Bio} +2.6\%), while also boosting the overall geometric accuracy to a peak AP of \textbf{90.0\%}. Furthermore, we initially hypothesized that applying a frequency-based weight to the regression objective would improve the localization of rare residual keypoints. However, empirical results indicate that scaling the regression loss actually degrades geometric performance, causing the AP to drop back to 89.7\%. This suggests that spatial coordinate regression is highly sensitive to artificial gradient scaling. Consequently, we discard regression weighting and apply distribution-aligned re-weighting strictly to the semantic classification branch.

\paragraph{Bio-Contrastive Boundary Loss ($\mathcal{L}_{bio}$) and Training Strategy.}
The inclusion of $\mathcal{L}_{bio}$ provides a strict logical constraint against anatomically impossible predictions. We compare two training strategies for incorporating this loss. Jointly training the entire model from scratch with $\mathcal{L}_{bio}$ forces the network to heavily prioritize semantic exclusivity, which slightly destabilizes spatial coordinate regression (AP drops to 89.6\%). Conversely, adopting a two-stage finetuning strategy, where $\mathcal{L}_{bio}$ is applied only after the model has acquired stable spatial features, yields the optimal results presented in Table \ref{tab:comprehensive_ablation}. This strategy retains peak geometric performance (\textbf{AP 90.0\%}) while dramatically boosting \texttt{Abs} accuracy to \textbf{85.6\%} and \texttt{Res-Bio} accuracy to \textbf{86.9\%} (a massive +5.8\% improvement over the weighted baseline). This strongly demonstrates that $\mathcal{L}_{bio}$ acts as a robust kinematic filter, successfully localizing the rare true amputation boundaries.

\paragraph{Effect of Synthetic Data and ProLoss.}
As shown in Table~\ref{tab:abl_syn_proloss}, adding synthetic data improves the recognition of rare semantic categories while maintaining comparable localization accuracy. For ViTPose-B, Real+Syn improves \texttt{Pros} accuracy from 55.1 to 63.3 and \texttt{Abs} accuracy from 56.6 to 58.5. A similar trend is observed for RTMPose-M, where \texttt{Pros} and \texttt{Abs} accuracies increase from 63.5/67.3 to 79.8/76.0. Incorporating ProLoss further improves the challenging categories, especially residual keypoints and \texttt{Pros}/\texttt{Abs} semantic types, while keeping AP nearly unchanged. These results demonstrate that the performance gains come from both the data expansion and the structure-aware objective.

\begin{table*}[t]
\centering
\caption{Ablation study on synthetic data and ProLoss.}
\label{tab:abl_syn_proloss}
\setlength{\tabcolsep}{3.2pt}
\small
\begin{tabular}{lccc|ccccc}
\toprule
Backbone & Data & Loss & AP & Type-Acc & Bio & Pros & Abs & Residual \\
\midrule
ViTPose-B & Real & Standard & 82.0 & 93.2 & 98.2 & 55.1 & 56.6 & 39.1 \\
ViTPose-B & Real+Syn & Standard & 81.9 & 93.6 & 97.8 & 63.3 & 58.5 & 40.7 \\
ViTPose-B & Real+Syn & ProLoss & 82.0 & 95.0 & 97.5 & \textbf{75.5} & \textbf{75.8} & \textbf{64.0} \\
\midrule
RTMPose-M & Real & Standard & 88.9 & 94.5 & 98.6 & 63.5 & 67.3 & 47.1 \\
RTMPose-M & Real+Syn & Standard & 88.7 & 96.1 & 98.7 & 79.8 & 76.0 & 67.9 \\
RTMPose-M & Real+Syn & ProLoss & 88.8 & 95.8 & 97.8 & \textbf{82.5} & \textbf{80.3} & \textbf{77.4} \\
\bottomrule
\end{tabular}
\end{table*}
\paragraph{Fine-grained Impact of Real-to-Synthetic Data Expansion.}
To further analyze the effectiveness of our Real-to-Synthetic data expansion pipeline, we conduct a fine-grained evaluation using the ViT-Base architecture. As shown in Table~\ref{tab:gen_improvement}, we compare the model trained exclusively on real-world data (Real Only) with the model trained on our enriched dataset (Real + Syn), and report both global semantic accuracy and per-joint prosthetic classification accuracy.

The results show that the synthetic data is particularly beneficial for long-tailed prosthetic joints. The global classification accuracy for Prosthetic (Pros) joints improves from 55.1\% to 63.3\%, while the accuracy for Absent (Abs) joints improves from 56.6\% to 58.5\%. More importantly, the gains are more pronounced on rare prosthetic categories. Specifically, the prosthetic classification accuracy for elbows increases from 30.2\% to 50.4\%, yielding a 20.2\% improvement. Wrists also show a substantial gain, increasing from 55.6\% to 70.2\%. Distal extremities and lower-limb joints, such as fingertips and knees, also improve by 7.6\% and 6.1\%, respectively.

These fine-grained results indicate that the Real-to-Synthetic data expansion pipeline effectively enriches rare prosthetic configurations and mitigates the long-tail bias in prosthetic joint recognition. Meanwhile, the coordinate regression performance remains comparable, with AP changing from 82.0\% to 81.9\% and AR improving from 81.0\% to 81.3\%. This suggests that the synthetic data improves semantic recognition of rare prosthetic joints without compromising the core spatial localization capability.

\begin{table}[!t]
    \caption{Impact of the Real-to-Synthetic data expansion on semantic classification and spatial localization. The enriched dataset mitigates the long-tail bias for prosthetic (Pros) and absent (Abs) joints, while maintaining comparable coordinate regression performance. We further report per-joint prosthetic classification accuracy for representative long-tail categories.}
    \label{tab:gen_improvement}
    \centering
    \resizebox{\columnwidth}{!}{
    \begin{tabular}{l|c|c|c}
        \toprule
        \textbf{Semantic Class} & \textbf{Real Only (Acc)} & \textbf{Real + Syn (Acc)} & \textbf{$\Delta$} \\
        \midrule
        Global: Pros ($\uparrow$) & 55.1\% & 63.3\% & \textbf{+8.2\%} \\
        Global: Abs ($\uparrow$) & 56.6\% & 58.5\% & \textbf{+1.9\%} \\
        \midrule
        Elbows (Pros) ($\uparrow$) & 30.2\% & 50.4\% & \textbf{+20.2\%} \\
        Wrists (Pros) ($\uparrow$) & 55.6\% & 70.2\% & \textbf{+14.6\%} \\
        Fingertips (Pros) ($\uparrow$) & 32.4\% & 40.0\% & \textbf{+7.6\%} \\
        Knees (Pros) ($\uparrow$) & 50.5\% & 56.6\% & \textbf{+6.1\%} \\
        \midrule
        \textbf{Coord. Reg.} & \textbf{Real Only} & \textbf{Real + Syn} & \textbf{$\Delta$}\\
        \midrule
        AP ($\uparrow$) & 82.0\% & 81.9\% & \textbf{-0.1\%} \\
        AR ($\uparrow$) & 81.0\% & 81.3\% & \textbf{+0.3\%} \\
        \bottomrule
    \end{tabular}}
\end{table}

\paragraph{Additional Comparative Analysis of Learning Objectives.}
Due to space constraints in the original conference manuscript, the supplementary material presented additional comparisons with standard coordinate regression and the existing LDLoss. A primary concern when introducing semantic classification into pose estimation is the potential degradation of spatial regression accuracy. However, compared to the ``Coordinate Regression Only'' baseline, equipping the model with ProLoss does not compromise the Average Precision (AP). We also observe that directly applying LDLoss to our dataset yields suboptimal geometric performance, because LDLoss is fundamentally designed under the assumption that the kinematic chain terminates at the residual stump. While effective for unprosthetized amputations, this constraint conflicts with our Omni-Pose protocol, which aims to reconstruct the entire mechanical structure of prostheses. In contrast, ProLoss adapts to the Omni-Pose protocol by extending the network's capability to perform semantic categorization while maintaining spatial AP.

\section{Conclusion}
We introduce \textbf{ProPose}, a comprehensive benchmark designed to resolve the representation bias and topological ambiguity in pose estimation. By proposing the \textbf{Omni-Pose} protocol, we effectively unify the semantic and spatial representation of intact, residual, and prosthetic limbs within a single framework.
Moreover, our real-to-synthetic pipeline effectively tackles data scarcity. By enforcing intra-limb semantic dependencies, \textbf{ProLoss} suppresses anatomically impossible predictions. Extensive evaluations show our approach significantly improves accuracy on specialized mechanical structures and residual boundaries. We hope ProPose advances inclusive human pose estimation.

\section*{Acknowledgments}
This research is funded in part by ARC-Discovery grant 
(DP220100800 to XY), ARC-DECRA grant (DE230100477 to XY), the Advance 
Queensland Industry Research Projects (AQIRP), and Follow ME PTY LTD. 
We thank all anonymous reviewers and ACs for their constructive suggestions.

\bibliographystyle{IEEEtran}
\bibliography{main}

\end{document}